\documentclass[sigconf]{acmart}
\AtBeginDocument{%
  \providecommand\BibTeX{{%
    \normalfont B\kern-0.5em{\scshape i\kern-0.25em b}\kern-0.8em\TeX}}}

\copyrightyear{2025}
\acmYear{2025}
\setcopyright{acmlicensed}\acmConference[KDD '25]{Proceedings of the 31st ACM SIGKDD Conference on Knowledge Discovery and Data Mining V.1}{August 3--7, 2025}{Toronto, ON, Canada}
\acmBooktitle{Proceedings of the 31st ACM SIGKDD Conference on Knowledge Discovery and Data Mining V.1 (KDD '25), August 3--7, 2025, Toronto, ON, Canada}
\acmDOI{10.1145/3690624.3709326}
\acmISBN{979-8-4007-1245-6/25/08}

\usepackage{multirow}
\usepackage{amsmath}
\usepackage{mathrsfs}
\usepackage{mathtools}
\usepackage{booktabs}
\usepackage{tabularx}
\usepackage{xcolor}
\usepackage{array}
\usepackage{colortbl}

\begin{document}

\title{LH-Mix: Local Hierarchy Correlation Guided Mixup over Hierarchical Prompt Tuning}

\author{Fanshuang Kong}
\email{kongfs@buaa.edu.cn}
\affiliation{
  \institution{CCSE, Beihang University}
  \city{Beijing}
  \country{China}
}

\author{Richong Zhang}
\authornote{Corresbonding author: zhangrc@act.buaa.edu.cn.}
\email{zhangrc@act.buaa.edu.cn}
\affiliation{
  \institution{CCSE, Beihang University}
  \city{Beijing}
  \country{China}
}

\author{Ziqiao Wang}
\email{ziqiaowang@tongji.edu.cn}
\affiliation{
  \institution{Tongji University}
  \city{Shanghai}
  \country{China}
}

\renewcommand{\shortauthors}{Fanshuang Kong, Richong Zhang, and Ziqiao Wang}

\begin{abstract}
  Hierarchical text classification (HTC) aims to assign one or more labels in the hierarchy for each text. Many methods represent this structure as a global hierarchy, leading to redundant graph structures.  To address this, incorporating a text-specific local hierarchy is essential. However, existing approaches often model this local hierarchy as a sequence, focusing on explicit parent-child relationships while ignoring implicit correlations among sibling/peer relationships. In this paper, we first integrate local hierarchies into a manual depth-level prompt to capture parent-child relationships. We then apply Mixup to this hierarchical prompt tuning scheme to improve the latent correlation within sibling/peer relationships. Notably, we propose a novel Mixup ratio guided by local hierarchy correlation to effectively capture intrinsic correlations. This \textbf{L}ocal \textbf{H}ierarchy \textbf{Mix}up (LH-Mix) model demonstrates remarkable performance across three widely-used datasets.
\end{abstract}



\maketitle

\section{Introduction}

\begin{figure}[ht]
    \centering
    \includegraphics[width=\columnwidth]{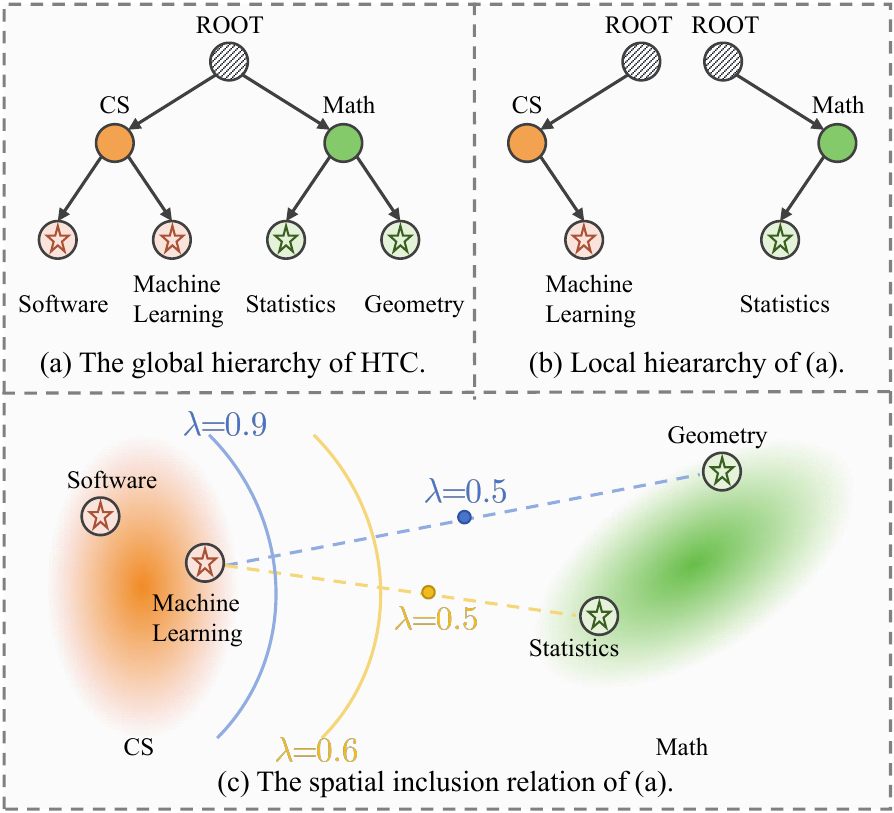}
    \caption{(a) A toy example of global hierarchy in HTC. (b) The local hierarchy of ``CS/Machine Learning'' and ``Math/Statistics'', which are extracted from (a). (c) Transformation from explicit parent-child relations (a) to spatial inclusion relations in latent space. Mixup enables the capture of varying degrees of implicit sibling/peer label correlation through different Mixup ratios $\lambda$.
    }
    \label{fig:motivation}
\end{figure}

Hierarchical Text Classification (HTC) is a variant of multi-label classification task characterized by labels organized in a predefined hierarchical structure \cite{silla2011survey}. This hierarchical structure captures relationships and dependencies among labels, with higher-depth labels containing lower-depth ones.  Each text is then assigned to one or more labels within this hierarchy.

The key challenge of HTC lies in effectively modeling the large-scale, imbalanced, and structured label hierarchy \cite{Mao_Tian_Han_Ren_2019}. 
Some existing works \cite{HTCInfoMax,HiAGM,HGCLR} consider the hierarchy as a directed acyclic graph, utilizing structure encoders to obtain label representations that incorporate hierarchical information. 
Nevertheless, this strategy allows each text to share the entire static global hierarchy, introducing redundancy in the graph \cite{hbgl}.
This redundancy becomes particularly noticeable as the hierarchy size increases, with numerous labels becoming irrelevant to a specific given target text \cite{im2023hierarchical}.
In contrast, alternative approaches proposed in \cite{hbgl,im2023hierarchical} invoke a local hierarchy, defined as a text-relevant sub-hierarchy extracted from the global hierarchy. Figure~\ref{fig:motivation} (a-b) illustrate examples of these two types of hierarchies.
This local hierarchy can be represented as an inclusion relationship in the latent space, as depicted in Figure~\ref{fig:motivation} (c). For example, ``Software'' and ``Machine Learning'' should occupy the same subspace within ``CS'', while ``Geometry'' and ``Statistics'' should exist in a subspace within the category of ``Math''.
On top of that, by treating a local hierarchy as a sequence, language models are able to capture parent-child relationships inherent in the hierarchy. Furthermore, motivated by the effectiveness of prompts in leveraging sequential information, in this work, we formulate the local hierarchy as a sequence using a hierarchical template to encode and align its structure.

It is worth mentioning that labels may hold inherent relevance that extends beyond the constraints imposed by parent-child relationships in the hierarchy. Figure~\ref{fig:motivation} (c) highlights a correlation between different local hierarchies. Specifically,  while ``CS/Machine Learning'' and ``Math/Statistics'' occupy  separate subspaces in the hierarchy, they may exhibit close proximity in the latent space, indicating a nuanced relationship that goes beyond the immediate hierarchical constraints. In fact, due to the widespread correlation among labels in HTC, including both sibling relationships and peer relationships \cite{song2023peer}, capturing the correlation between local hierarchies becomes crucial. Existing methods, however, lack explicit mechanisms to tackle this specific challenge. An avenue of promise lies in Mixup \cite{mixup}, a technique utilized for augmenting latent correlations between input pairs through the generation of intermediate samples \cite{mixup}. Therefore, alongside hierarchical prompt tuning, we incorporate Mixup to reveal and leverage the implicit correlation between local hierarchies.

More specifically, we first apply prompt tuning to HTC, where the local hierarchy is treated as a sequence. Virtual tokens are introduced to represent each depth in the hierarchy, thereby disassembling the parent-child relationship and aligning the hierarchy by depth. Within this hierarchical prompt tuning framework, we incorporate Mixup to capture implicit correlations among sibling and peer relationships. Notably, unlike Vanilla Mixup, which interpolates between the input and corresponding target using the same Mixup ratio sampled from a Beta Distribution, recent studies \cite{sawhney2022dmix, zhang2022m} adopt Mixup ratios based on instance correlations. This approach adjusts the mixture degree based on instance similarity, assigning higher mixture ratios to more similar instances to generate more informative samples. Conversely, less similar examples invoke a lower degree of mixture to prevent the generation of out-of-distribution data \cite{baena2022preventing}. As shown in Figure \ref{fig:motivation} (c), ``CS/Machine Learning'' should have a higher degree of mixing (e.g., with a Mixup ratio $0.6$) with ``Math/Statistic'', and a lower degree of mixing (e.g., with a Mixup ratio $0.9$) to ``Math/Geometry''. Motivated by this observation, we propose a novel approach for Mixup ratio control in our framework, dubbed local hierarchy Mixup (LH-Mix). In particular, LH-Mix assesses the correlation of a local hierarchy pair using the representation derived from the hierarchical prompt. By incorporating a heuristic function to determine the correlation and Mixup ratio, LH-Mix is able to learn hierarchical label correlations more effectively than vanilla Mixup.

Below we summarize the main contributions of our work:
\begin{itemize}
    \item We introduce LH-Mix, which integrates Mixup based on a depth-level hierarchical prompt, effectively modeling the local hierarchy within sequences. To the best of our knowledge, LH-Mix is the first application of Mixup in HTC.
    \item To capture the inherent correlations among local hierarchies, we propose a strategy for adjusting the Mixup ratio. This adaptive approach, guided by the local hierarchy correlation, encourages the generation of enriched in-between samples.
    \item We evaluate our LH-Mix on three standard HTC datasets, demonstrating its effectiveness through the empirical performance. 
\end{itemize}

\section{Related Work}
\subsection{Hierarchical Text Classification (HTC)}
HTC refers to a specific type of multi-label text classification problem, where an instance lies in one or more paths from a taxonomic hierarchy in a top-down manner. 
The hierarchical structure is typically represented as a tree or a directed acyclic graph, where the root represents the highest-level label, and the leaf nodes correspond to the most specific labels.
The key challenge in HTC lies in how to effectively utilize the
large-scale, imbalanced, and structured label hierarchy. 
To perform HTC, various techniques can be employed.
Previous works can be categorized into two types \cite{HiAGM,HTCInfoMax}: global approach and local approach.
The global approach tackles the HTC problem as a flat multi-label classification problem, employing global hierarchy as input and building a single classifier for all labels. There are various strategies that incorporate the global hierarchy, such as capsule network \cite{Peng_Li_Gong_Wang_He_Li_Wang_Yu_2019}, reinforcement learning \cite{Mao_Tian_Han_Ren_2019}, meta-learning \cite{Wu_Xiong_Wang_2019}, contrastive learning \cite{HGCLR}, hyperbolic representation \cite{chen2020hyperbolic}, or structure encoder \cite{HiAGM,HTCInfoMax}. 
On the contrary, the local approach typically involves constructing individual classifiers for each label \cite{Banerjee_Akkaya_Perez-Sorrosal_Tsioutsiouliklis_2019}, each parent node \cite{Dumais_Chen_2000}, or each level of the label hierarchy \cite{Shimura_Li_Fukumoto_2018,Wehrmann_Barros_Cerri_2018,kowsari2017hdltex}. 

The current SOTA HTC model, HBGL \cite{hbgl}, proposes hierarchy-guided BERT with both global and local hierarchies to utilize the prior knowledge of the pre-trained language model. The suboptimal model HPT \cite{HPT} transforms the global prediction to a local one by hierarchical prompt tuning. These successful applications of local hierarchy have served as inspiration for further investigation of local hierarchy in this paper.

\subsection{Mixup}
Mixup is a data augmentation method proposed by \cite{mixup}, which aims to enhance the generalization capabilities of neural networks by generating in-between samples through linearly interpolating pairs of input text and their corresponding labels. Vanilla Mixup interpolates input pairs by a mixing ratio sampling from a basic Beta Distribution.
However, the mixing ratio obtained through a blind sampling process may not be optimal. To enhance performance, several studies have attempted to control the mixing ratio for pairs. AdaMixUp \cite{guo2019mixup} incorporates the automatic mixing policies from the data through the utilization of an additional network and an objective function to prevent ``manifold intrusion''. CAMixup \cite{wen2020combining} adjusts the mixing ratio through the relation between the predict confidence and accuracy. Remix \cite{chou2020remix} assigns labels in a manner that favors the minority class by providing disproportionately higher weights to the minority class. Nonlinear Mixup \cite{guo2020nonlinear} incorporates a nonlinear interpolation policy for both the input and label pairs, wherein the mixing policy for the labels is adaptively learned, leveraging the information from the mixed input. Basically, Mixup could enhance model performance by an adaptive mixing ratio that captures deeper relationships within the latent space between samples.

Due to the strong scalability and effectiveness, Mixup and its variants have shown promising improvements in various tasks, such as classification \cite{guoNLPmixup,CutMix}, unsupervised domain adaptation \cite{mao_ma_yang_chen_li_2019,wu_inkpen_el-roby_2020}, and semi-supervised learning \cite{mixtext}.
However, in the context of HTC, there has been limited exploration of applying Mixup to augment the hierarchical label representation.

\begin{figure*}[ht]
    \centering
    \includegraphics[width=0.975\linewidth]{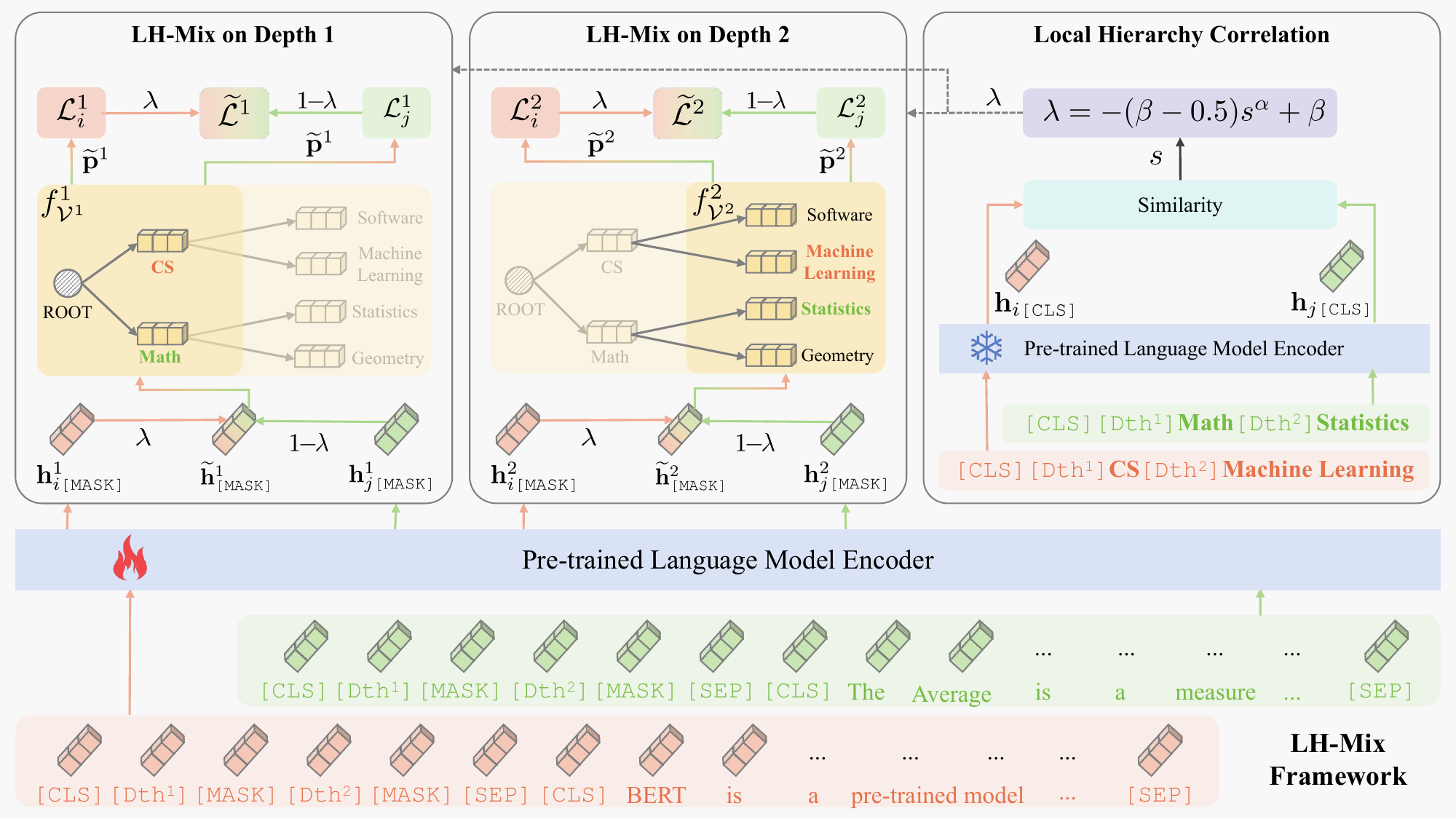}
    \caption{Illustration of LH-Mix. The light orange color scheme represents elements of $X_i$, and the light green represents $X_j$. The mixture of orange and green represents the elements related to the Mixup operation. 
    }
    \label{fig:model}
\end{figure*}

\section{Preliminaries}

\subsection{HTC Setting}
Given a training dataset $\{(X_i, Y_i)\}_{i=1}^N$, where $X_i$ denotes the input text, $Y_i$ is the corresponding label of $X_i$, and $N$ denotes the size of training dataset.
Let $\mathcal{V}$ be the label set. In the HTC setting, $\mathcal{V}$ is further divided into $D$ different subsets, denoted as $\mathcal{V}=\{\mathcal{V}^1, ..., \mathcal{V}^d, ..., \mathcal{V}^D \}$, where $D$ is the total depth of the hierarchy and $\mathcal{V}^d$ is the label set of the $d$-th depth.
Importantly, $\mathcal{V}$ can be organized into a tree structure, representing the global hierarchy of the entire dataset. In addition, each input $X_i$ contains multiple labels from $\mathcal{V}$, and these labels can form a separate subtree, known as the local hierarchy.
The goal of HTC is to assign labels from the local hierarchy in $\mathcal{V}$ to each text in the test dataset.

\subsection{Hierarchical Prompt for HTC}
Typically, a prompt contains a task-specific template with the text serving as input, exploiting the knowledge ingrained within the pre-trained language model \cite{brown2020language}. 
A vanilla prompt framework in HTC, as outlined in \cite{HPT}, involves the creation of $D$ soft tokens within the template, each specifically associated with a depth in the hierarchy. For instance, the input can be formulated as:

\texttt{[CLS]} \texttt{[Dth$^1$]} \texttt{[MASK]} ... \texttt{[Dth$^d$]} \texttt{[MASK]} ... \texttt{[Dth$^D$]} \texttt{[MASK]} \texttt{[SEP]}

$X$ \texttt{[SEP]}

\noindent In particular, each token \texttt{[Dth}$^d$\texttt{]} serves to prompt the prediction of the associated label $\mathcal{V}^d$ within its $d$-th layer hierarchy. This utilizes the hidden output of the \texttt{[MASK]} token immediately following it. 
More precisely, let $\mathbf{h}_\texttt{[MASK]}^d$ be the hidden output of \texttt{[MASK]} in the $d$-th layer, the classification procedure is then defined as:
\begin{equation}
\mathbf{p}^d=f_{\mathcal{V}^d}^d(\mathbf{h}_{\texttt{[MASK]}}^d),
\end{equation}
where $\mathbf{p}^d \in \mathbb{R}^{|\mathcal{V}^d|}$ is the prediction score vector, and $f_{\mathcal{V}^d}^d$ is a mapping from hidden output to the prediction scores in the $d$-th layer. 
Note that $f_{\mathcal{V}^d}^d$ contains a classifier trained through the Masked Language Model task, along with a label words verbalizer that adheres to the label words of $\mathcal{V}^d$.
Notably, by using this hierarchical prompt for HTC,
the classification process is now partitioned across each level of the hierarchy, rather than predicting all labels in a single classifier. This approach encodes the local hierarchy as a sequence, allowing for alignment of the hierarchy at each depth level for every input. Consequently, this hierarchical prompt design greatly facilitates subsequent Mixup procedures. We will elaborate on this later.

In the HTC task, many existing models treat the task as a multiple-label classification and invoke the conventional Binary Cross Entropy (BCE) loss. Nevertheless, it has been pointed out that BCE loss ignores the correlation between labels. A more effective alternative, as suggested in \cite{HPT}, is the Zero-bounded Multi-label Cross Entropy (ZMLCE). Particularly, the ZMLCE loss considers that all scores for positive labels are greater than $0$, while those for negative labels are less than $0$. The ZMLCE loss is formally defined as follows:
\begin{equation}
        \mathcal{L}^d (\mathbf{p}^d, \mathcal{V}^d_{pos}, \mathcal{V}^d_{neg}) = \log (1+\sum_{v \in \mathcal{V}^d_{neg}}e^{\mathbf{p}^d_{\cdot v}}) + \log (1+\sum_{v \in \mathcal{V}^d_{pos}}e^{-\mathbf{p}^d_{\cdot v}}),
\end{equation}
where $\mathcal{V}^d_{pos}$ and $\mathcal{V}^d_{neg}$ are the positive label set and negative label set in the $d$-th layer, respectively, and $\cdot v$ denotes the $v$-th element of a vector (e.g., $\mathbf{p}^d_{\cdot v}$ refers to the $v$-th element of the prediction score vector). In the inference phase, a prediction score greater than $0$ is considered as positive; otherwise, it is considered as negative.

\section{LH-Mix Framework}
Mixup, as introduced in \cite{mixup}, uses linear interpolation between inputs and their respective labels to generate in-between samples, thereby enriching the inherent structures within the latent space \cite{wu_inkpen_el-roby_2020}. 
In this work, we integrate Mixup into the hierarchical prompt framework to better capture the correlations within the hierarchically structured labels of HTC.
In particular, we propose a novel adaptive Mixup ratio strategy, guided by the local hierarchy correlation, to further refine the vanilla Mixup scheme. We term this enhanced approach LH-Mix.
The framework of LH-Mix is shown in Figure \ref{fig:model}.

\subsection{Local Hierarchy Correlation}
The design of this hierarchical prompt scheme is motivated by the fact that the local hierarchy can be effectively represented as a sequence. Specifically, an important characteristic of hierarchy is its strict hierarchical relationship, where the prediction of lower-level labels relies on the prediction of higher-level labels.
To achieve this characteristic, we introduce a depth token \texttt{[Dth}$^d$\texttt{]} to represent the depth level of each label.
The local hierarchy can be reformulated to a sequence of tokens and this will be further translated to a soft token as hierarchical prompts via a pre-trained language model. We believe this soft prompt integrates both the level and the label information, allowing the representation of hierarchies as sequences.
More specially, for a specific input $X$, its local hierarchy can similarly be expressed as such a sequence by substituting the \texttt{[MASK]} token with the corresponding gold label. 
For example, the two local hierarchies ``CS/Machine Learning'' and ``Math/Statistics'' in Figure \ref{fig:motivation} can be respectively represented as:

$\texttt{[CLS]}  \texttt{[Dth$^1$]}  \text{CS} \texttt{[Dth$^2$]} \text{Machine Learning} \texttt{[SEP]}$

$\texttt{[CLS]} \texttt{[Dth$^1$]} \text{Math} \texttt{[Dth$^2$]} \text{Statistics} \texttt{[SEP]}$

\noindent Consequently, we transform the local hierarchy into a ``sentence''. In addition, when a local hierarchy contains multiple paths, we concatenate the labels at the same level together and place them after the corresponding \texttt{[Dth$^d$]} token. 

Furthermore, notice that by feeding a sentence into a pre-trained language model and extracting the hidden output of \texttt{[CLS]} token, one can obtain a sentence embedding that can be used to calculate sentence similarity \cite{reimers2019sentence}. Similarly, in our context, the \texttt{[CLS]} output corresponding to the ``local hierarchy sequence'' serves as the representation of the local hierarchy. This representation can be effectively utilized to compute the local hierarchy correlation.
Formally, consider two inputs $X_i$ and $X_j$, let their local hierarchy representations be denoted by $\mathbf{h}_{i\texttt{[CLS]}}$ and $\mathbf{h}_{j\texttt{[CLS]}}$, respectively. The distance between these two hierarchies is measured by the normalized cosine similarity, namely,
\begin{equation}
    \begin{aligned}
        s & = 0.5\left(\frac{\mathbf{h}_{i\texttt{[CLS]}} \cdot \mathbf{h}_{j\texttt{[CLS]}}}{\|\mathbf{h}_{i\texttt{[CLS]}}\| \|\mathbf{h}_{j\texttt{[CLS]}}\|} + 1\right), 
    \end{aligned}
\end{equation}
where $\cdot$ denotes the vector dot product and $\|\cdot\|$ is the $L^2$ norm. Note that due to the shared hierarchical encoding format, local hierarchy representations use the same pre-trained model encoder. Nonetheless, it is important to highlight that the encoder utilized in similarity calculation, distinct from the one used for classification, remains unaffected by gradient updates.

\subsection{Local Hierarchy Correlation Guided Mixup Ratio}

\begin{figure}[ht]
    \centering
    \includegraphics[width=\columnwidth]{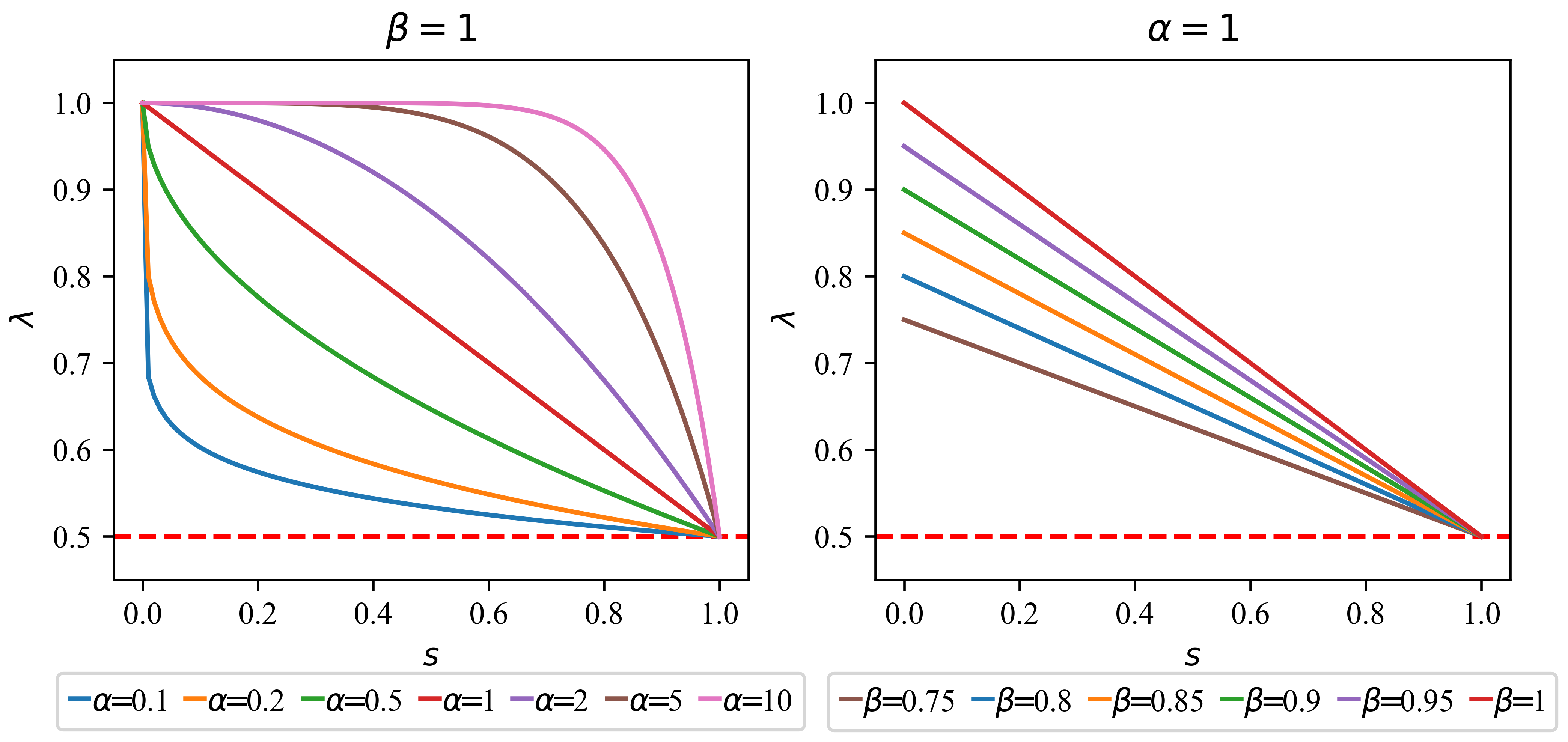}
    \caption{Curves corresponding to Eq. \ref{eq:s_lambda}. We separately plot the effects of different $\beta$ and $\alpha$ on the function when $\alpha=1$ and $\beta=1$.}
    \label{fig:s_lambda}
\end{figure}

Mixup captures sample correlation by generating in-between samples via a Mixup ratio. The magnitude of the Mixup ratio can be controlled to generate varying degrees of ``hard examples'', thereby determining the extent to which Mixup affects the data \cite{chou2020remix,zhang2022unsupervised}. \cite{sawhney2022dmix,baena2022preventing,zhang2022m} demonstrate that examples with different levels of similarity should be mixed with varying intensities.
Therefore, given the varying correlation among different local hierarchy pairs, it is more appropriate in this context to apply different Mixup ratios to each local hierarchy pair, rather than drawing the Mixup ratio from a fixed distribution (e.g., Beta distribution).
While numerous studies have explored the relationship between correlation and the Mixup ratio, there is no well-established theoretical framework that precisely characterizes the numerical relationship between them. In fact, rigorously justifying the effectiveness of Mixup still remains an open problem. Hence, to reflect the relationship between the Mixup ratio and similarity, we design a heuristic function based on intuition.
The underlying intuition here is that as the similarity between two local hierarchies increases, we expect Mixup to better capture the latent correlation between them. Specifically, we expect Mixup to have more impact (i.e., with a Mixup ratio approaching $0.5$) on highly correlated local hierarchies, whereas its impact is diminished (i.e., with a Mixup ratio approaching $1$) on less correlated local hierarchies.
Bear this in mind, to formulate the relationship between local hierarchy similarity $s$ and Mixup ratio $\lambda$, we have heuristically designed the following function:
\begin{equation}
    \lambda = -(\beta - 0.5)s^\alpha + \beta,
\label{eq:s_lambda}
\end{equation}
where $\alpha > 0$ controls the rate of change of $\lambda$ with respect to $s$, and $\beta \in (0.5, 1]$ controls the upper bound of $\lambda$. 
This function has the advantage of covering various common linear or nonlinear relationships between $s$ and $\lambda$ by simply adjusting the values of $\alpha$ and $\beta$.
To gain better understanding about Eq.~\ref{eq:s_lambda}, we visualize the effects of $\beta$ and $\alpha$ in Figure~\ref{fig:s_lambda}. In particular, when $\alpha=1$, there is a linear relationship between $s$ and $\lambda$. When $\alpha<1$, $\lambda$ decreases at a slower rate as $s$ increases. Conversely, when $\alpha>1$, $\lambda$ decreases at a faster rate as $s$ increases. Moreover, $\beta$ determines the maximum value of $\lambda$, signifying the minimum impact of Mixup.

\subsection{Local Hierarchy Mixup}
We now apply Mixup with the local hierarchy correlation guided ratio to both input and output simultaneously at each depth of the hierarchy.

Regarding input Mixup, following the approach of previous Mixup variants in text classification \cite{guoNLPmixup}, we interpolate the hidden output corresponding to the \texttt{[MASK]} token for an input pair at each depth of the hierarchy as follows:
\begin{equation}
    \begin{aligned}
    \widetilde{\mathbf{h}}^d_{\texttt{[MASK]}} = \lambda \mathbf{h}^d_{i\texttt{[MASK]}} + (1 - \lambda) \mathbf{h}^d_{j\texttt{[MASK]}}.
    \end{aligned}
\end{equation}
Subsequently, we obtain the prediction score $\widetilde{\mathbf{p}}^d$ for the mixed input $\widetilde{\mathbf{h}}^d_{\texttt{[MASK]}}$ using $f_{\mathcal{V}^d}^d$, namely,
\begin{equation}
    \widetilde{\mathbf{p}}^d = f_{\mathcal{V}^d}^d(\widetilde{\mathbf{h}}^d_{\texttt{[MASK]}}).
\end{equation}

For output Mixup, the straightforward approach involves mixing the labels, as formulated in the vanilla Mixup technique
\cite{mixup}. Alternatively, several studies have demonstrated that mixing the losses is also a valid method.
In particular, it has been established that gradients for label mixing and loss mixing are equivalent in the context of cross-entropy loss \cite{chang2021single}.
In the scenario of the ZMLCE loss, characterized by the division of positive and negative texts with $0$ as an anchor, 
all positive labels and negative labels are treated as distinct combinations.
While this design enables ZMLCE to focus on correlations between labels, it neglects the consideration of relative magnitudes among positive labels or negative labels. This oversight can obscure the interpretation of label mixing, making its meaning less explicit.
Therefore, we opt to mix the loss terms in our LH-Mix, defined as:
\begin{equation}
    \begin{aligned}
        &  \widetilde{\mathcal{L}}^d (\lambda, \widetilde{\mathbf{p}}^d, \mathcal{V}^d_{i, pos}, \mathcal{V}^d_{i, neg}, \mathcal{V}^d_{j, pos}, \mathcal{V}^d_{j, neg})\\
        = &  \lambda\mathcal{L}^d_i (\widetilde{\mathbf{p}}^d, \mathcal{V}^d_{i, pos}, \mathcal{V}^d_{i, neg}) + (1 - \lambda) \mathcal{L}^d_j (\widetilde{\mathbf{p}}^d, \mathcal{V}^d_{j, pos}, \mathcal{V}^d_{j, neg}).
    \end{aligned}
\label{eq:loss}
\end{equation}

Further, the gradient of models with parameters $\theta$ is given by:

\begin{equation}
    \begin{aligned}
        & \nabla_\theta \biggl(\lambda \biggl(\log (1+\sum_{v \in \mathcal{V}^d_{i, neg}}e^{ \widetilde{\mathbf{p}}^d_{\cdot v}}) + \log (1+\sum_{v \in \mathcal{V}^d_{i, pos}}e^{-\widetilde{\mathbf{p}}^d_{\cdot v}}) \biggr)\\
        & +  (1 - \lambda) \biggl(\log (1+\sum_{v \in \mathcal{V}^d_{j, neg}}e^{\widetilde{\mathbf{p}}^d_{\cdot v}}) + \log (1+\sum_{v \in \mathcal{V}^d_{j, pos}}e^{-\widetilde{\mathbf{p}}^d_{\cdot v}}) \biggr) \biggr) \\
        = & \lambda(\frac{\sum_{v \in \mathcal{V}^d_{i, neg}}e^{\widetilde{\mathbf{p}}^d_{\cdot v}}}{1 + \sum_{v \in \mathcal{V}^d_{i, neg}}e^{\widetilde{\mathbf{p}}^d_{\cdot v}}} - \frac{\sum_{v \in \mathcal{V}^d_{i, pos}}e^{-\widetilde{\mathbf{p}}^d_{\cdot v}}}{1 + \sum_{v \in \mathcal{V}^d_{i, pos}}e^{-\widetilde{\mathbf{p}}^d_{\cdot v}}}) \\
         & + (1 - \lambda)(\frac{\sum_{v \in \mathcal{V}^d_{j, neg}}e^{\widetilde{\mathbf{p}}^d_{\cdot v}}}{1 + \sum_{v \in \mathcal{V}^d_{j, neg}}e^{\widetilde{\mathbf{p}}^d_{\cdot v}}} - \frac{\sum_{v \in \mathcal{V}^d_{j, pos}}e^{-\widetilde{\mathbf{p}}^d_{\cdot v}}}{1 + \sum_{v \in \mathcal{V}^d_{j, pos}}e^{-\widetilde{\mathbf{p}}^d_{\cdot v}}})
    \end{aligned}
    \label{eq:lossmix}
\end{equation}

Clearly, 
Eq.~\ref{eq:lossmix} delivers a linear combination for the gradients of the mixed prediction scores corresponding to  $X_i$ and $X_j$.
In addition, the vanilla Mixup, specifically in the case of cross-entropy, also generates a linear combination of the gradients from the mixed prediction score. In this sense, the loss mixing for ZMLCE aligns more closely with the vanilla Mixup. This also validates our decision to adopt Eq.~\ref{eq:loss} for optimization, favoring loss mixing over label mixing.

\section{Experiment}

\subsection{Datasets and Evaluation Metrics}
We evaluate our model on three widely used datasets: WebOfScience (WOS) \cite{kowsari2017hdltex}, NYTimes (NYT) \cite{Sandhaus_2008} and RCV1-V2 \cite{Lewis_Yang_Rose_Li_2004}. The statistic information is shown in Table \ref{tab:data}.
For evaluation metrics, we adopt Macro-F1 and Micro-F1 to measure the results following the previous work \cite{HPT,hbgl,he2023instances,HGCLR}. 
Micro-F1 considers the overall precision and recall of all instances, while Macro-F1 represents the average F1-score across labels.

\begin{table}[ht]
    \centering
    \begin{tabular}{c|cccccc}
        \toprule
        Dataset & $D$ & $|\mathcal{V}|$ & Avg($|\mathcal{V}_i|$) & \#Train & \#Dev & \#Test \\
        \midrule
        WOS & 2 & 141 & 2.0 & 30,070 & 7,518 & 9,397 \\
        NYT & 8 & 166 & 7.6 & 23,345 & 5,834 & 7,292 \\
        RCV1-V2 & 4 & 103 & 3.24 & 20,833 & 2,316 & 781,265 \\
        \bottomrule
    \end{tabular}
    \caption{Data statistics. $D$ is the maximum depth of the hierarchy. $|\mathcal{V}|$ is the number of labels. Avg($|\mathcal{V}_i|$) is the average number of labels per instance. \# represents the number of instance in the datasets.}
    \label{tab:data}
\end{table}

\begin{table*}[ht]
    \centering
    \begin{tabular}{ccccccc}
        \toprule
        \multirow{2}{*}{\textbf{Model}} & \multicolumn{2}{c}{\textbf{WOS}} & \multicolumn{2}{c}{\textbf{NYT}} & \multicolumn{2}{c}{\textbf{RCV1-V2}} \\
        \cline{2-7}
         & Micro-F1 & Macro-F1 & Micro-F1 & Macro-F1 & Micro-F1 & Macro-F1\\
         \midrule
         \multicolumn{7}{c}{\textbf{Hierarchy-aware Models}} \\
         TextRNN \cite{HiAGM} & 83.55 & 76.99 & 70.83 & 56.18 & 81.57 & 59.25 \\
         HiAGM \cite{HiAGM} & 85.82 & 80.28 & 74.97 & 60.83 & 83.96 & 63.35 \\
         HTCInfoMax \cite{HTCInfoMax} & 85.58 & 80.05 & - & - & 83.51 & 62.71 \\
         HiMatch \cite{HiMatch} & 86.20 & 80.53 & - & - & 84.73 & 64.11 \\
         \midrule
         \multicolumn{7}{c}{\textbf{Large Language Model}} \\
         ChatGPT \cite{he2023instances} & - & - & - & - & 51.35 & 32.20 \\
         LLaMA & 83.36 & 73.34 & 76.72 & 58.01 & 87.49 & 62.12 \\
         \midrule
         \multicolumn{7}{c}{\textbf{Pre-trained Language Model}} \\
         BERT \cite{HGCLR} & 85.63 & 79.07 & 78.24 & 65.62 & 85.65 & 67.02 \\
         BERT+HiAGM \cite{HGCLR} & 86.04 & 80.19 & 78.64 & 66.76 & 85.58 & 67.93 \\
         BERT+HTCInfoMax \cite{HGCLR} & 86.30 & 79.97 & 78.75 & 67.31 & 85.53 & 67.09 \\
         BERT+HiMatch \cite{HiMatch} & 86.70 & 81.06 & - & - & 86.33 & 68.66 \\
         HGCLR \cite{HGCLR} & 87.11 & 81.20 & 78.86 & 67.96 & 86.49 & 68.31 \\
         HPT \cite{HPT} &  87.16 & 81.93 & 80.42 & \underline{70.42} & \underline{87.26} & 69.53 \\
         HBGL \cite{hbgl} & \underline{\textbf{87.36}} & \underline{82.00} & 80.47 & 70.19 & 87.23 & \underline{71.07} \\
         HJCL \cite{he2023instances} & - & - & \underline{80.52} & 70.02 & 87.04 & 70.49 \\
         HiTIN \cite{hitin} & 87.19 & 81.57 & 79.65 & 69.31 & 86.71 & 69.95 \\
         \rowcolor{gray!15} LH-Mix & 87.27 & \textbf{82.07} & \textbf{80.57} & \textbf{71.26} & \textbf{87.47} & \textbf{71.93} \\
        \bottomrule
    \end{tabular}
    \caption{Main results of LH-Mix and other baseline models. The underlined results are the current state-of-the-art (SOTA). The bolded results are the best performance including LH-Mix.}
    \label{tab:main-result}
\end{table*}

\begin{table*}[ht]
    \centering
    \begin{tabular}{ccccccc}
        \toprule
        \multirow{2}{*}{\textbf{Model}} & \multicolumn{2}{c}{\textbf{WOS}} & \multicolumn{2}{c}{\textbf{NYT}} & \multicolumn{2}{c}{\textbf{RCV1-V2}} \\
        \cline{2-7}
         & Micro-F1 & Macro-F1 & Micro-F1 & Macro-F1 & Micro-F1 & Macro-F1\\
         \midrule
        Bert & 85.63 & 79.07 & 78.24 & 65.62 & 85.65 & 67.02 \\
        +Mixup & 86.58 & 81.05 & 78.89 & 67.85 & 86.36 & 67.33 \\
        +LH-Mix & \textbf{86.86} & \textbf{81.31} & \textbf{79.17} & \textbf{68.46} & \textbf{86.52} & \textbf{70.04} \\
         \midrule
         Prompt & 86.94 & 81.37 & 80.23 & 70.13 & 87.11 & 69.82 \\
         + Mixup & 87.13 & 81.82 & 80.35 & 70.63 & 87.45 & 71.39 \\
         + LH-Mix & \textbf{87.27} & \textbf{82.07} & \textbf{80.57} & \textbf{71.26} & \textbf{87.47} & \textbf{71.93}\\
        \bottomrule
    \end{tabular}
    \caption{Ablation study of different variants.}
    \label{tab:ablation}
\end{table*}

\subsection{Implement Details}
Following the previous works, we exploit the pre-trained model \texttt{bert-base-uncased} from Hugging Face Transformers~\footnote{\url{https://github.com/huggingface/transformers}} to evaluate our model. For hierarchical prompting, the newly added $D$ tokens \texttt{[Dth$^d$]} are randomly initialized, and the label words verbalizer embeddings are initialized by the average of the label name representation. All parameters are fine-tuned by an Adam~\cite{kingma2014adam} optimizer with the learning rate as 3e-5.
For LH-Mix related parameters,
as Figure \ref{fig:s_lambda} shows, we choose $\alpha$ from $[0.1, 0.3, 0.6, 1, 2, 5, 10]$ and $\beta$ from $[0.7, 0.75, 0.8, 0.85, 0.9, 0.95, 1]$.
This configuration is able to encompass the majority of common relationships between $s$ and $\lambda$.
To accelerate the convergence of the model, we adopt a two-step training strategy. Initially, we train the model for 5 epochs without utilizing Mixup, and then for the remaining epochs, we incorporate Mixup during the training process. We employ the early stopping strategy if there is no improvement of Macro-F1 after 5 epochs. 
All experiments are conducted on a Tesla V100-SXM2-32GB GPU, with each epoch taking approximately 1800s for training on medium size dataset NYT. 
For more details about training, we will release code our the Github~\footnote{https://github.com/fskong/LH-Mix}.

\subsection{Comparable Models}
As the practice in previous works, we compare LH-Mix with three groups of models: hierarchy-aware model, large language model, and pre-trained language model. In the hierarchy-aware model, we compare our model with 4 strong baselines: TextRNN \cite{HiAGM}, HiAGM \cite{HiAGM}, HTCInfoMax \cite{HTCInfoMax}, and HiMatch \cite{HiMatch}. In the large language model, we generally report instruction-tuned results by ChatGPT \cite{chatgpt} and supervised fine-tuned results of LLaMA-2-7B \cite{touvron2023llama}. In the pre-trained language model, except for substituting the encoder as BERT of the 4 mentioned hierarchy-aware baselines, we also consider 5 newly proposed models including HGCLR \cite{HGCLR}, HPT \cite{HPT}, HBGL \cite{hbgl}, HJCL \cite{he2023instances}, and HiTIN \cite{hitin}. Among these baseline models, HPT and HBGL overall achieve state-of-the-art performance. For more details on these comparable models, please refer to Appendix \ref{sec:details}.

\begin{table*}[ht]
    \centering
    \begin{tabular}{ccccccc}
        \toprule
        \multirow{2}{*}{\textbf{Model}} & \multicolumn{2}{c}{\textbf{WOS}} & \multicolumn{2}{c}{\textbf{NYT}} & \multicolumn{2}{c}{\textbf{RCV1-V2}} \\
        \cline{2-7}
         & Micro-F1 & Macro-F1 & Micro-F1 & Macro-F1 & Micro-F1 & Macro-F1\\
         \midrule
        Prompt & \textbf{86.94} & \textbf{81.37} & 80.23 & 70.13 & 87.11 & 69.82 \\
        \midrule
        +GCN & 86.81 & 81.31 & 80.28 & 70.45 & \textbf{87.22} & 68.63 \\
        +GAT & 86.91 & 81.20 & \textbf{80.30} & \textbf{70.65} & 87.16 & 69.95 \\
        +Graphormer & 86.86 & 81.03 & 80.21 & 69.79 & 87.17 & \textbf{70.02} \\
        \bottomrule
    \end{tabular}
    \caption{Comparison with GNN-based Models.}
    \label{tab:gnn-based}
\end{table*}

\begin{table*}[ht]
    \centering
    \begin{tabular}{ccccccc}
        \toprule
        \multirow{2}{*}{\textbf{Model}} & \multicolumn{2}{c}{\textbf{WOS}} & \multicolumn{2}{c}{\textbf{NYT}} & \multicolumn{2}{c}{\textbf{RCV1-V2}} \\
        \cline{2-7}
         & Micro-F1 & Macro-F1 & Micro-F1 & Macro-F1 & Micro-F1 & Macro-F1\\
         \midrule
        Prompt & 86.94 & 81.37 & 80.23 & 70.13 & 87.11 & 69.82 \\
        \midrule
        +EDA & 86.64 & 81.24 & 79.73 & 69.34 & 86.98 & 69.22 \\
        +BT & 86.63 & 80.98 & 79.77 & 69.13 & 87.00 & 70.75 \\
        +LH-Mix & \textbf{87.27} & \textbf{82.07} & \textbf{80.57} & \textbf{71.26} & \textbf{87.47} & \textbf{71.93} \\
        \bottomrule
    \end{tabular}
    \caption{Comparison with data augmentation methods.}
    \label{tab:data augmentation}
\end{table*}

\subsection{Main Results}
\label{main results}
The main results of Micro-F1 and Macro-F1 of three datasets are shown in Table \ref{tab:main-result}. As the Table shows, LH-Mix achieves the best performance in five out of the total six metrics, indicating the effectiveness of LH-Mix. To further evaluate the reproducibility and significance of LH-Mix, we conduct experiments on statistical performance in Appendix \ref{sec:statistical performance}.

With a detailed investigation of these results, we first observe that WOS shows a smaller improvement compared to NYT and RCV1-V2. We believe this is due to the smaller number of the depth of WOS, which results in easier classification. 
We then observe that LH-Mix shows a larger improvement in Macro-F1.
We believe this is because Macro-F1 measures label-level F1, while Micro-F1 measures more strict instance-level F1. LH-Mix primarily improves the hierarchical label correlation, thus leading to a greater improvement for Macro-F1.

Furthermore, comparing the HTC results of the instruction-tuned large language model reported in \cite{he2023instances}, which is based on ChatGPT \texttt{gpt-turbo-3.5}, the results indicate that large models still face significant challenges in encoding complex hierarchical structures. Comparing the results with the supervised fine-tuned large language models, like LLaMA, we found that fully fine-tuning BERT-based models achieve better performance, indicating the encoder architecture is still effective in natural language understanding. Also, considering the training time for LLaMA is significantly longer than BERT, we believe that LH-Mix currently has distinctive advantages.

\subsection{Ablation Study}
To evaluate the importance of each variant in LH-Mix, we conduct a series of ablation experiments, including the following variations: BERT, which serves as the baseline model without any additional variant; Prompt, which introduces hierarchical templates for prompts tuning; +Mixup, which apply the vanilla Mixup with Mixup ratio from conventional Beta Distribution based on BERT or Prompt; +LH-Mix, which further utilize hierarchy correlation guided Mixup ratio for +Mixup.
The results are shown in Table \ref{tab:ablation}.

The results that Prompt is superior to the basic BERT model, demonstrating the effectiveness of hierarchical templates. Additionally, the performance of the +Mixup is better than its corresponding basic BERT or Prompt model, indicating the utility of using Mixup for encoding hierarchical label correlation. Furthermore, +LH-Mix outperforms +Mixup, providing evidence for the capability of incorporating hierarchical label correlation in Mixup ratio controlling.

\subsection{Comparison with GNN-based Models}
\label{sec:gnn-based}

We observe that numerous studies have suggested that graph encoders can obtain better hierarchical label embeddings. However, graph encoders primarily focus on modeling the global hierarchy. As outlined in our introduction, we believe that capturing the local hierarchy can also provide sufficient information.
Thus, we compare Prompt with a series of GNN-based graph models, such as GCN~\cite{kipf2016semi}, GAT~\cite{velickovic2017graph} and Graphormer~\cite{ying2021transformers}, to evaluate the effect of the hierarchical prompt scheme in Table~\ref{tab:gnn-based}.

From the Table, we find that the gain from the graph encoder is limited. Considering the extra time cost, we believe the hierarchical prompt is effective in capturing hierarchical information.

\subsection{Comparison with Data Augmentation Methods}
\label{sec:data aug}
As LH-Mixup is a data augmentation method, we evaluated two other widely-used data augmentation techniques for text—Easy Data Augmentation (EDA) and Back Translation (BT). EDA employs simple operations such as synonym replacement, random insertion, deletion, or swapping to generate augmented samples. In contrast, BT leverages machine translation by translating text into another language and back to create variations while preserving semantics.

The results, shown in Table~\ref{tab:data augmentation}, demonstrate that both EDA and BT perform worse compared to LH-Mix. This comparison highlights the effectiveness of LH-Mix as a text data augmentation approach.

\begin{figure}[ht]
    \centering
    \begin{tabular}{c}
    WOS \\
    \includegraphics[width=\columnwidth]{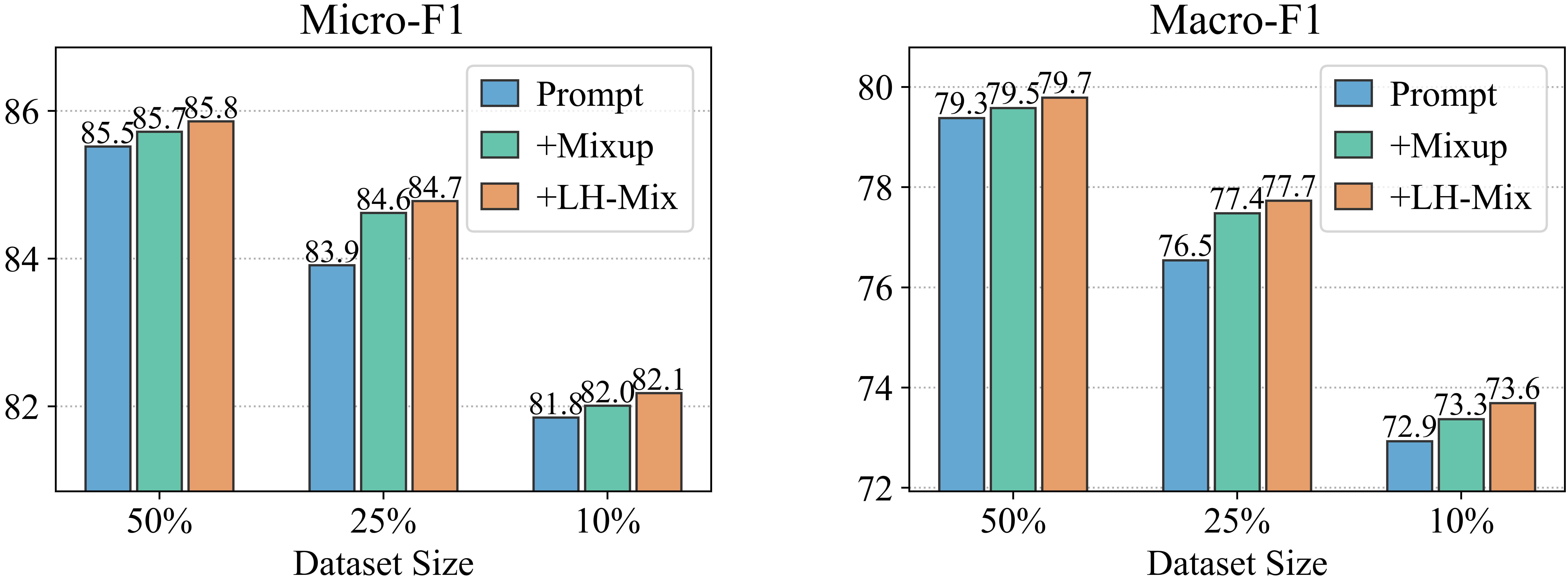}\\
    NYT \\
    \includegraphics[width=\columnwidth]{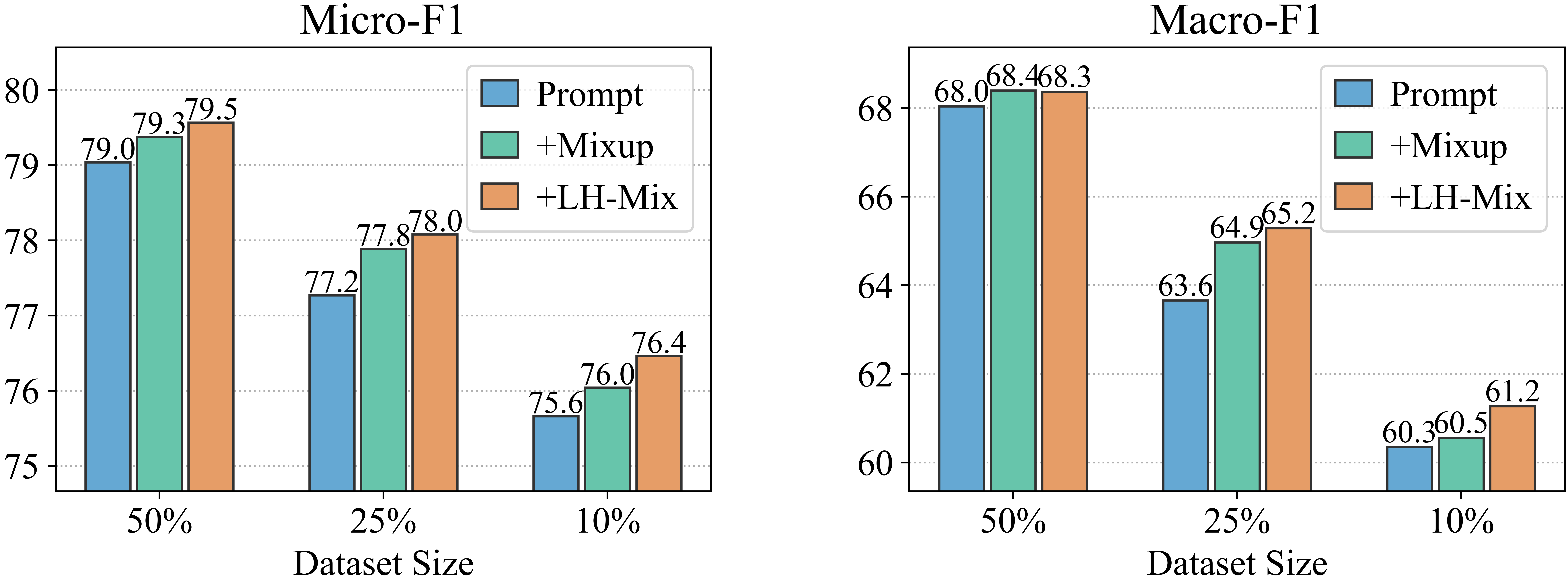}\\
    RCV1-V2 \\
    \includegraphics[width=\columnwidth]{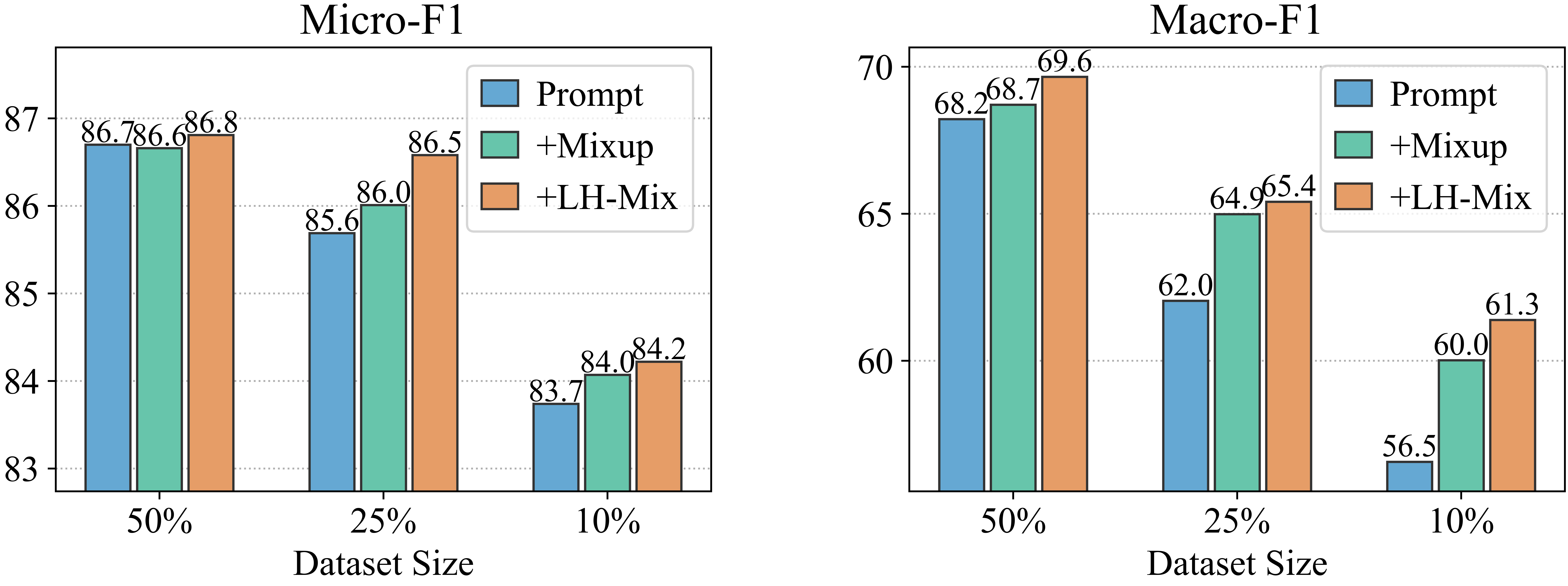}
    \end{tabular}
    \caption{Performance on different downsample ratios to sparse data.}
    \label{fig:sparse}
\end{figure}

\subsection{Performance on Sparse Datasets}
Since LH-Mix is essentially a regularization method, it is necessary to evaluate LH-Mix in the low-resource setting. Therefore, we downsample $50\%$, $25\%$, $10\%$ of the training data for analysis. The results of Prompt, vanilla Mixup (+Mixup) and LH-Mix (+LH-Mix) are shown in Figure \ref{fig:sparse}.

Results in the Figure indicate that +Mixup and +LH-Mix continue to perform better than Prompt with the decreasing of training samples. This also proves the effectiveness of Mixup methods in hierarchy label correlation capturing on even sparser datasets.
Additionally, +LH-Mix is better than +Mixup generally, and as the dataset decreases extremely to 10\%, the performance gap between +LH-Mix and +Mixup is enlarged.
This observation demonstrates the efficiency of the adaptive Mixup ratio guided by local hierarchy correlation.

\begin{figure}[ht]
    \centering
    \begin{tabular}{c}
        $\beta=1$ \\
        \includegraphics[width=\columnwidth]{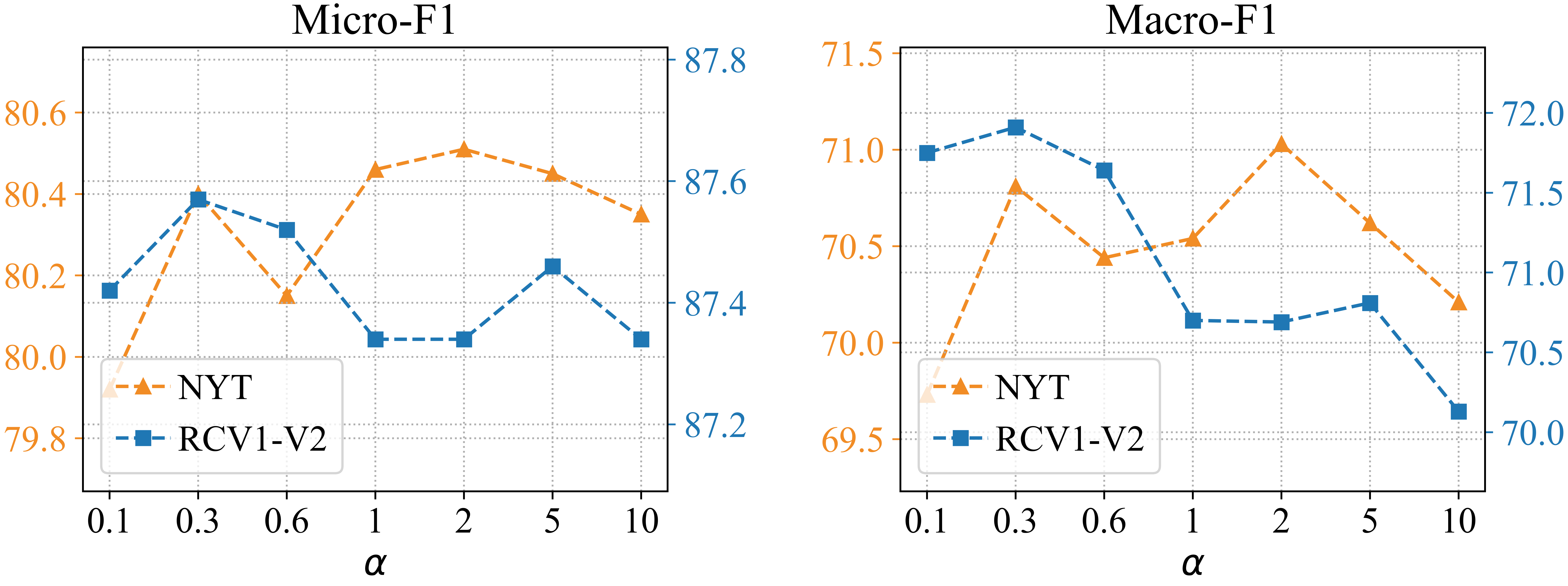} \\
        $\alpha=1$ \\
        \includegraphics[width=\columnwidth]{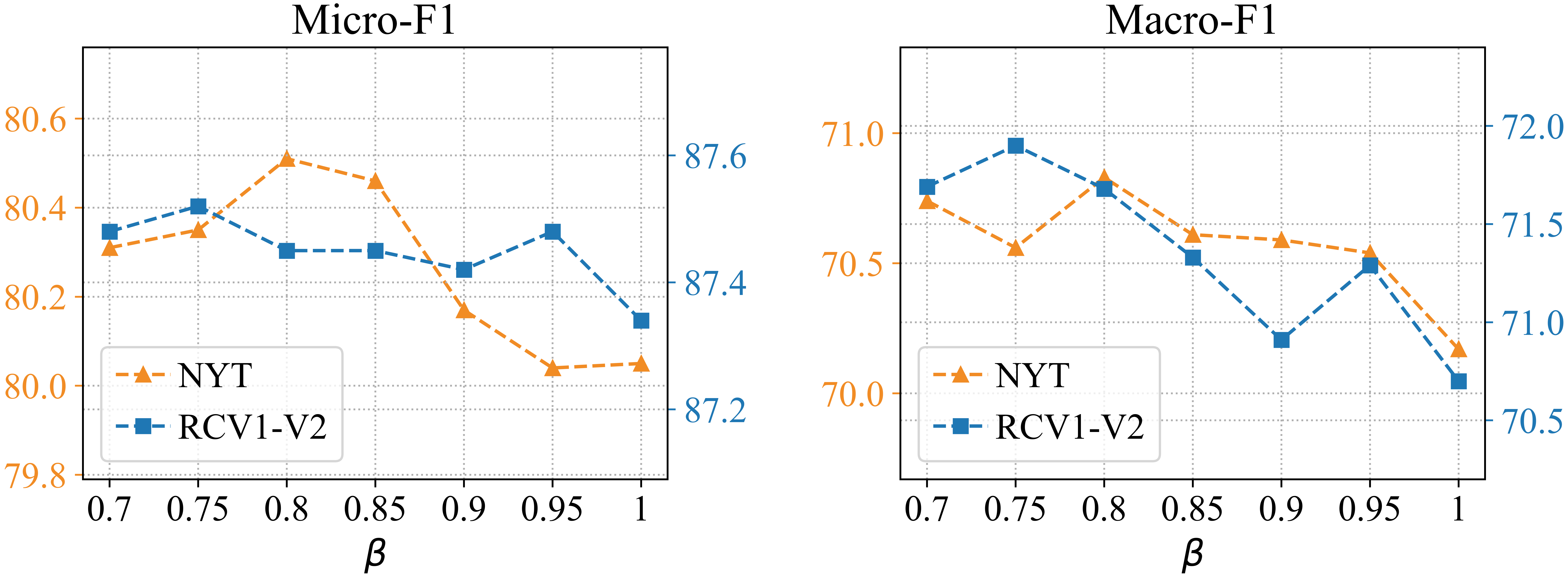} \\
         
    \end{tabular}
    \caption{Performance on different $\beta$ and $\alpha$, when fixing $\alpha=1$ and $\beta=1$ respectively.}
    \label{fig:parameter}
\end{figure}

\begin{figure*}[ht]
    \centering
    \begin{tabular}{m{0.275\linewidth} c m{0.6\linewidth}}
  \includegraphics[width=\linewidth]{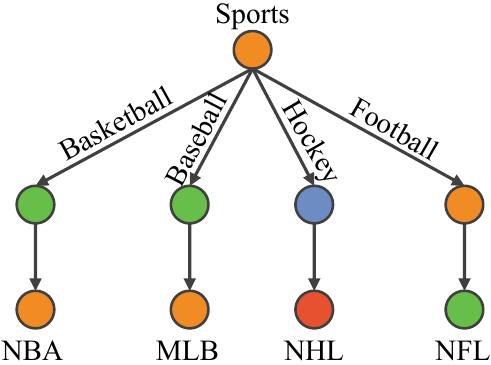} & \quad\quad\quad & \includegraphics[width=\linewidth]{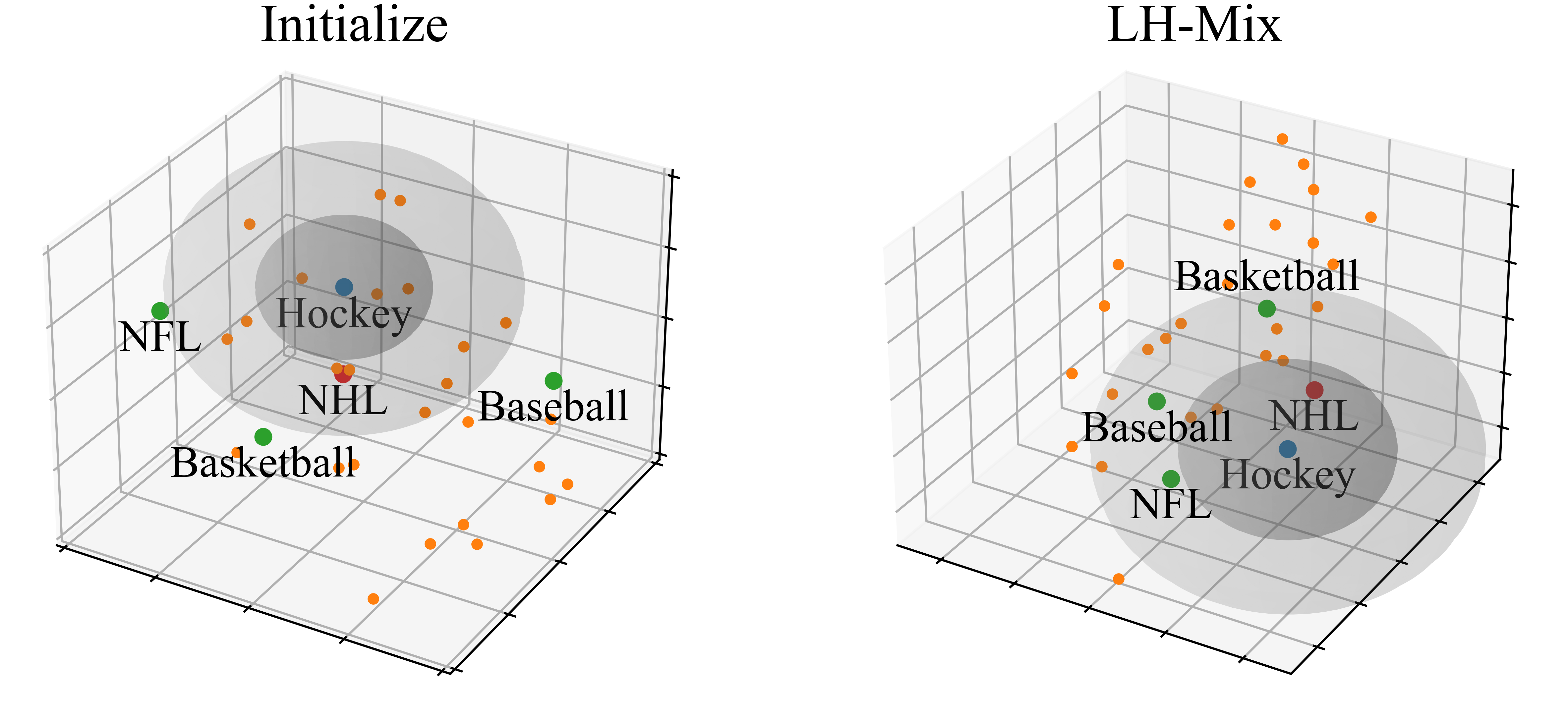} \\
  \centering (a) Related labels of ``Hockey'' & \quad\quad\quad & \centering  (b) Visualization of label correlation
\end{tabular}
    \caption{(a) Example of hierarchically related labels on ``Hockey''. (b) The visualization of label correlations related to ``Hockey'' (the blue dot) before and after training by LH-Mix. We select the most similar 30 labels to ``Hockey'' for display and mark them with orange scatters. Two shades of gray spheres construct the top 3 and 15 similar label spaces, respectively. Red dots indicate labels that are not initially ranked within the top 3 but are included after training, while green dots represent labels that are not initially ranked within the top 15 but are included after training.}
    \label{fig:tsne}
\end{figure*}

\subsection{Parameter Analysis}
In LH-Mix, the parameters $\alpha$ and $\beta$ are crucial for controlling the relationship between hierarchy correlation and Mixup ratio.
To investigate the impact of $\alpha$ and $\beta$, we fix $\alpha=1$ and $\beta=1$, and analyze the results as the other parameter varied. The results are shown in Figure \ref{fig:parameter}. Due to the relatively small changes of Micro-F1 (as explained in Section \ref{main results}), the trends may not accurately reflect the final results. Therefore, we temporarily focus on the Macro-F1 for consideration.

\subsubsection{Effect of $\alpha$} When $\beta=1$ is fixed, for NYT, the Macro-F1 initially increases and then decreases with increasing $\alpha$. For RCV1-V2, the Macro-F1 gradually decreases. 
We speculate that this is likely due to differences in the statistical characteristics of the datasets. This demonstrates that the relationship between $s$ and $\lambda$ does indeed impact the effectiveness of LH-Mix.

\subsubsection{Effect of $\beta$}
Reviewing Figure \ref{fig:s_lambda}, as $\beta$ increases, there is a higher likelihood for $\lambda$ to be larger, indicating that LH-Mix is more likely to have a minor effect. From Figure \ref{fig:parameter}, 
when $\alpha=1$ is fixed, the results for both NYT and RCV1-V2 gradually decrease with increasing $\beta$.
This indicates that a higher degree of LH-Mix leads to better performance, confirming the effectiveness of the LH-Mix.

\begin{figure}[ht]
    \centering
    \begin{tabular}{c}
    \includegraphics[width=\columnwidth]{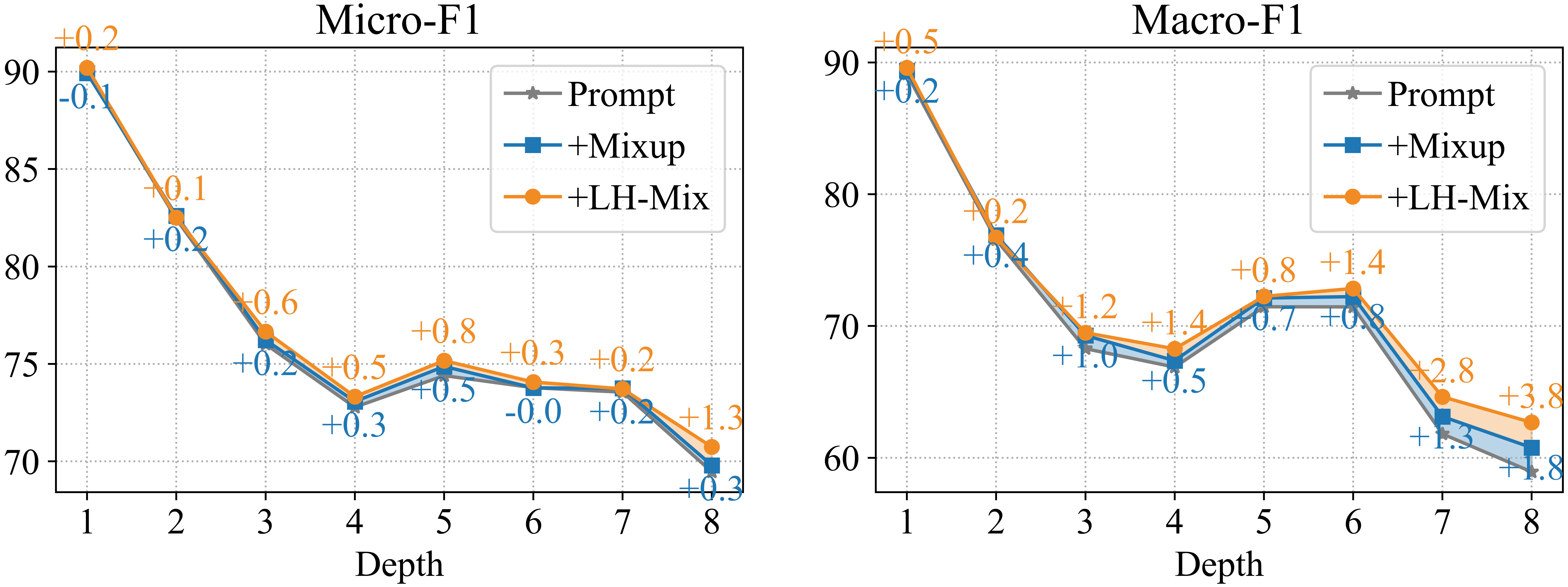}\\
    \includegraphics[width=\columnwidth]{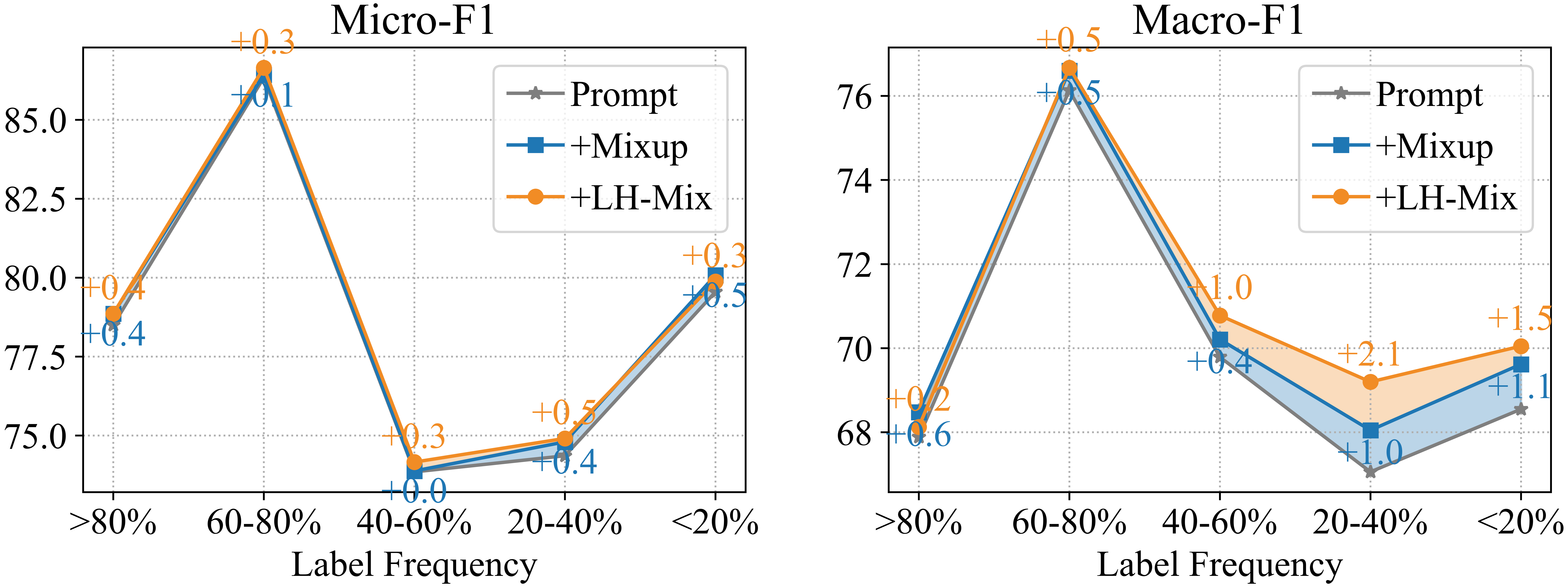}
    \end{tabular}
    \caption{Performance of comparable models on different depths of the hierarchy and different frequencies of labels. The texts in the Figure indicate the superior performance of +Mixup and +LH-Mix compared to Prompt.
    }
    \label{fig:hierarchy}
\end{figure}

\subsection{Variation of Label Correlation by LH-Mix}

To further analyze whether LH-Mix learns more accurate hierarchical label correlation, we conduct a case study on label similarity. Specifically, we select the label ``Hockey'' as the target and compute the top 30 similar labels before and after learning with LH-Mix. The visualization of label correlation by t-SNE \cite{van2008visualizing} is shown in Figure \ref{fig:tsne}, marked by Initialize and LH-Mix.

From the Figure, we first observe that the most related label ``NHL'' has a higher similarity rank after learning (from top 15 to top 3). Additionally, the rank of labels ``Baseball'', ``Basketball'', and ``NFL'', which are also closely related to ``NHL'', have improved (from top 30 to top 15). Noted, for conciseness and clarity, we only emphasize the labels with significant differences before and after training. However, for labels with minor variations, such as ``NBA'', ``MLB'', and ``Football'', we do not provide any special markings. This does not imply that these labels are irrelevant; rather, it is because these labels have consistently been learned effectively. 
These observations indicate better performance in hierarchical label correlation after learning with LH-Mix.

\subsection{Improvements of LH-Mix on Hierarchy}
To evaluate the performance of LH-Mix in different hierarchical structures, we analyze the performance of the Prompt, vanilla Mixup (+Mixup), and LH-Mix (+LH-Mix) models on the NYT dataset, which contains the most complex hierarchical structure. Specifically, the complexity of the hierarchical structure is reflected in two aspects: the depth of the hierarchy and the distribution of label frequency. It is generally believed that the deeper the label hierarchy and the lower the label frequency, the more challenging it is to learn label representations, resulting in lower accuracy. The comparison of these models is shown in Figure \ref{fig:hierarchy}.

From the Figure, we find that as the label hierarchy deepened and the label frequency decreased, the improvement of +Mixup and +LH-Mix compared to the baseline Prompt is larger and larger. This demonstrates the effectiveness of Mixup-related methods in enhancing the learning performance of labels in complex structures. Furthermore, we observe that the improvement achieved by +LH-Mix is superior to that of +Mixup, indicating that our proposed LH-Mix method can further capture the hierarchical label correlation.

\section{Conclusion}
HTC is an important scenario within multi-label text classification and has numerous applications. In this paper, we first propose a hierarchical template to model and align the local hierarchy in the prompt tuning framework. By employing this hierarchical prompt tuning, we formulate the parent-child relationships explicitly and effectively. Based on this, we employ Mixup to capture the implicit sibling/peer relationships under the latent label space. Especially we induce a local hierarchy correlation guided Mixup strategy to regulate the Mixup ratio for improved hierarchical label correlations, named LH-Mix. Extensive experiments on three widely-used HTC datasets confirm the effectiveness of our model.

\begin{acks}
This work was supported by the National Science and Technology Major Project under Grant 2022ZD0120202, in part by the National Natural Science Foundation of China (No. U23B2056), in part by the Fundamental Research Funds for the Central Universities, and in part by the State Key Laboratory of Complex \& Critical Software Environment.
\end{acks}

\bibliographystyle{ACM-Reference-Format}
\bibliography{custom}

\appendix
\begin{table*}[ht]
    \centering
    \begin{tabular}{ccccccc}
        \toprule
        \multirow{2}{*}{\textbf{Model}} & \multicolumn{2}{c}{\textbf{WOS}} & \multicolumn{2}{c}{\textbf{NYT}} & \multicolumn{2}{c}{\textbf{RCV1-V2}} \\
        \cline{2-7}
         & Micro-F1 & Macro-F1 & Micro-F1 & Macro-F1 & Micro-F1 & Macro-F1\\
         \midrule
         HPT & $86.84_{\pm0.12}$ & $81.29_{\pm0.10}$ & $80.28_{\pm0.16}$ & $70.02_{\pm0.25}$ & $87.19_{\pm0.05}$ & $69.50_{\pm0.12}$ \\
         HBGL & $87.18_{\pm0.14}$ & $81.69_{\pm0.26}$ & $80.28_{\pm0.14}$ & $69.78_{\pm0.14}$ & $86.97_{\pm0.14}$ & $70.54_{\pm0.42}$ \\
         LH-Mix & $87.15_{\pm0.14}$& $\textbf{81.82}_{\pm0.18}$ & $\mathbf{80.51}_{\pm±0.15}$ & $\mathbf{71.03}_{\pm0.24}$ & $\mathbf{87.39}_{\pm0.12}$ & $\mathbf{71.49}_{\pm0.50}$ \\
        \bottomrule
    \end{tabular}
    \caption{Statistical performance on SOTA models and LH-Mix.}
    \label{tab:stability}
\end{table*}

\begin{table*}[ht]
    \centering
    \begin{tabular}{ccccccc}
        \toprule
        \multirow{2}{*}{\textbf{P-value}} & \multicolumn{2}{c}{\textbf{WOS}} & \multicolumn{2}{c}{\textbf{NYT}} & \multicolumn{2}{c}{\textbf{RCV1-V2}} \\
        \cline{2-7}
         & Micro-F1 & Macro-F1 & Micro-F1 & Macro-F1 & Micro-F1 & Macro-F1\\
         \midrule
        LH-Mix vs. HPT & \textbf{0.0008} & \textbf{0.00001} & 0.063 & \textbf{0.0004} & \textbf{0.0032} & \textbf{0.00003} \\
        LH-Mix vs. HBGL & 0.4902 & 0.2733 & \textbf{0.011} & \textbf{0.000005} & \textbf{0.0060} & \textbf{0.0041} \\
        \bottomrule
    \end{tabular}
    \caption{T-test of LH-Mix compared to HPT and HBGL.}
    \label{tab:t-test}
\end{table*}

\section{Appendix}
\subsection{Details of Comparable Models}
\label{sec:details}

\noindent \textbf{TextRNN} utilizes TextRNN to encode the input and treat HTC as a global text classification problem.

\noindent \textbf{HiAGM} \cite{HiAGM} utilizes a directed graph hierarchy and hierarchy-aware structure encoders to model label dependencies. It proposes an end-to-end hierarchy-aware global model with two variants: HiAGM-LA, which learns hierarchy-aware label embeddings and performs inductive fusion of label-aware text features, and HiAGM-TP, which directly feeds text features into hierarchy encoders.

\noindent \textbf{HTCInfoMax} \cite{HTCInfoMax} proposes an information maximization approach with two modules: text-label mutual information maximization and label prior matching. The first module captures interactions between text and labels to filter irrelevant information, while the second enhances the structure encoder's ability to represent all labels, tackling label imbalance in HTC.

\noindent \textbf{HiMatch} \cite{HiAGM} introduces a hierarchy-aware label semantics matching network to formulate the text-label semantics relationship. It projects text and label semantics into a joint embedding space and utilizes a joint embedding loss and a matching learning loss to model the matching relationship.

\noindent \textbf{HGCLR} \cite{HGCLR} introduces hierarchy-guided contrastive learning, embedding the label hierarchy into the text encoder by constructing positive samples based on the hierarchy for hierarchy-aware text representation.

\noindent \textbf{HPT} \cite{HPT} introduces hierarchy-aware prompt tuning for HTC using multi-label MLM. It employs dynamic virtual templates and soft prompts with label words to integrate label hierarchy knowledge, alongside a zero-bounded multi-label cross-entropy loss to align HTC and MLM objectives.

\noindent \textbf{HBGL} \cite{hbgl} introduces Hierarchy-guided BERT with global and local hierarchies, which leverages large-scale parameters and prior language knowledge to model hierarchies. It directly models semantic and hierarchical information with BERT, avoiding the intentional fusion of separate modules.

\noindent \textbf{HJCL} \cite{he2023instances} introduces hierarchy-aware joint supervised contrastive learning, which combines supervised contrastive learning with HTC. It utilizes instance-wise and label-wise contrastive learning techniques and carefully constructs batches to achieve the contrastive learning objective.

\noindent \textbf{HiTIN} \cite{hitin} proposes hierarchy-aware tree isomorphism network to enhance text representations using only label hierarchy's syntactic information. It converts the label hierarchy into a coding tree guided by structural entropy and incorporates hierarchy-aware information into text representations through a structure encoder.

\noindent \textbf{ChatGPT} represents utilizing instruction-tuned language models, such as ChatGPT.
Results are directly reported from \cite{he2023instances}. They manually design templates and use \texttt{gpt-turbo-3.5} for predictions. 
Noted, \cite{he2023instances} does not report the results for WOS and NYT, which is probably because the flattened hierarchical labels in the template occupy most of the tokens. Texts and labels in WOS and NYT are relatively long, making them hard to implement.

\noindent \textbf{LLaMA} represents utilizing fine-tuned language models, such as LLaMA.
We conduct classification training using the hidden layer output of the last token of the LLaMA-2-7B. Additionally, we use LoRA to efficiently fine-tune the LLM.

\subsection{Statistical Performance of LH-Mix}
\label{sec:statistical performance}

\subsubsection{Performance on Reproducibility}
To ensure reproducibility and robustness, we repeat the experiments five times using different random seeds for LH-Mix and comparable models. For reasons of performance and code accessibility, we specially select the two SOTA models HPT and HBGL for comparison. The average and standard deviation results are reported in Table \ref{tab:stability}. From the Table, we find that all models exhibit relatively stable performance. Moreover, LH-Mix consistently achieves optimal performance than other models overall.

\subsubsection{T-test with SOTA Models}
We conduct T-tests on LH-Mix compared to the two SOTA models, HPT and HBGL. The resulting P-values are presented in Table \ref{tab:t-test}. As per the conventional practice, a P-value less than 0.05 indicates statistical significance. From our analysis, we observe that LH-Mix exhibits a statistically significant improvement over HPT in 5 out of the 6 metrics, and over HBGL in 4 out of 6 metrics. Combining these findings with the results discussed in Section \ref{main results} of our paper, we conclude that the improvement achieved by LH-Mix is statistically significant.

\end{document}